\documentclass[letterpaper, 10 pt, conference]{ieeeconf}  

\usepackage{multirow}
\usepackage{amsmath}    
\usepackage{graphicx}
\usepackage{hyperref}
\usepackage{subfigure}
\usepackage{color}
\usepackage{times} 

\title{\LARGE \bf
Visual Recognition of Isolated Swedish Sign Language Signs
}

\author{Saad Akram$^1$ ~~~~~~~ Jonas Beskow$^2$ ~~~~~~ Hedvig Kjellstr{\"o}m$^1$\\
         $^1$CVAP/CAS, KTH, Stockholm, Sweden  ~~~ $^2$Speech, Music and Hearing, KTH, Stockholm, Sweden\\
         {\tt saadua,beskow,hedvig@kth.se}}

\begin{document}

\maketitle
\thispagestyle{empty}
\pagestyle{empty}

\begin{abstract}
\label{sec:abstract}
We present a method for recognition of isolated Swedish Sign Language signs. The method will be used in a game intended to help children training signing at home, as a complement to training with a teacher. The target group is not primarily deaf children, but children with language disorders. Using sign language as a support in conversation has been shown to greatly stimulate the speech development of such children. 
The signer is captured with an RGB-D (Kinect) sensor, which has three advantages over a regular RGB camera. Firstly, it allows complex backgrounds to be removed easily. We segment the hands and face based on skin color and depth information. Secondly, it helps with the resolution of hand over face occlusion. Thirdly, signs take place in 3D; some aspects of the signs are defined by hand motion vertically to the image plane. This motion can be estimated if the depth is observable. The 3D motion of the hands relative to the torso are used as a cue together with the hand shape, and HMMs trained with this input are used for classification. 
To obtain higher robustness towards differences across signers, Fisher Linear Discriminant Analysis is used to find the combinations of features that are most descriptive for each sign, regardless of signer. 
Experiments show that the system can distinguish signs from a challenging 94 word vocabulary with a precision of up to 94\% in the signer dependent case and up to 47\% in the signer independent case.

\end{abstract}

\section{INTRODUCTION}
\label{sec:intro}

Automatic sign language recognition (SLR) is a challenging research topic that has gained interest rapidly during the past decade. The potential benefits of this technology are obvious: with an ever-increasing information flow in today's society, the sign-language speaking communities are often left to communicate in their second language -- the local verbal language. Functional systems for sign language recognition and translation would allow signers to communicate in their first language and be understood by non-signers. 

\begin{figure}[t]
\centerline{%
\subfigure[Two performances of 'Kan jag f\aa' ('Can I have')]{\includegraphics[width=0.24\textwidth]{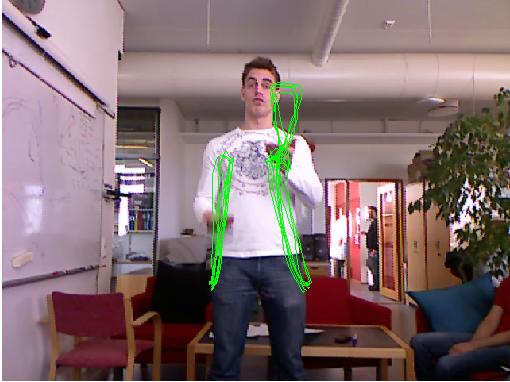}\hspace{0.2mm}
\includegraphics[width=0.24\textwidth]{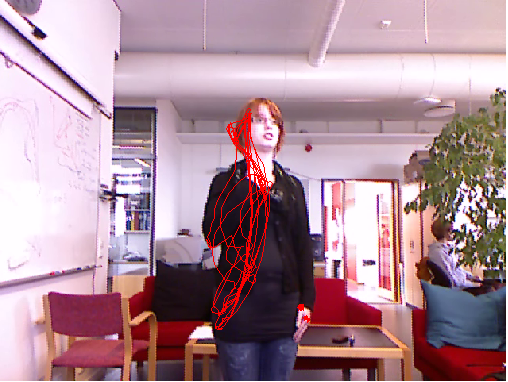}}}
\vspace{-2mm}
\centerline{%
\subfigure['R{\"a}dd' ('Scared')]{\includegraphics[width=0.24\textwidth]{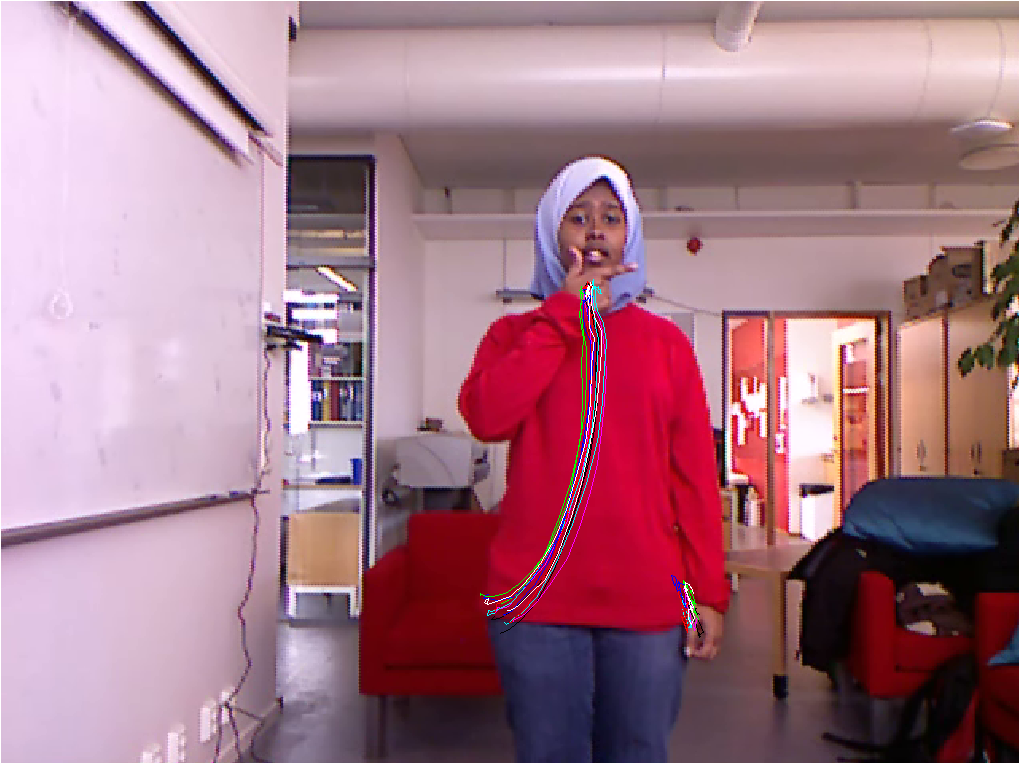}}  
\subfigure['Farlig' ('Dangerous')]{\includegraphics[width=0.24\textwidth]{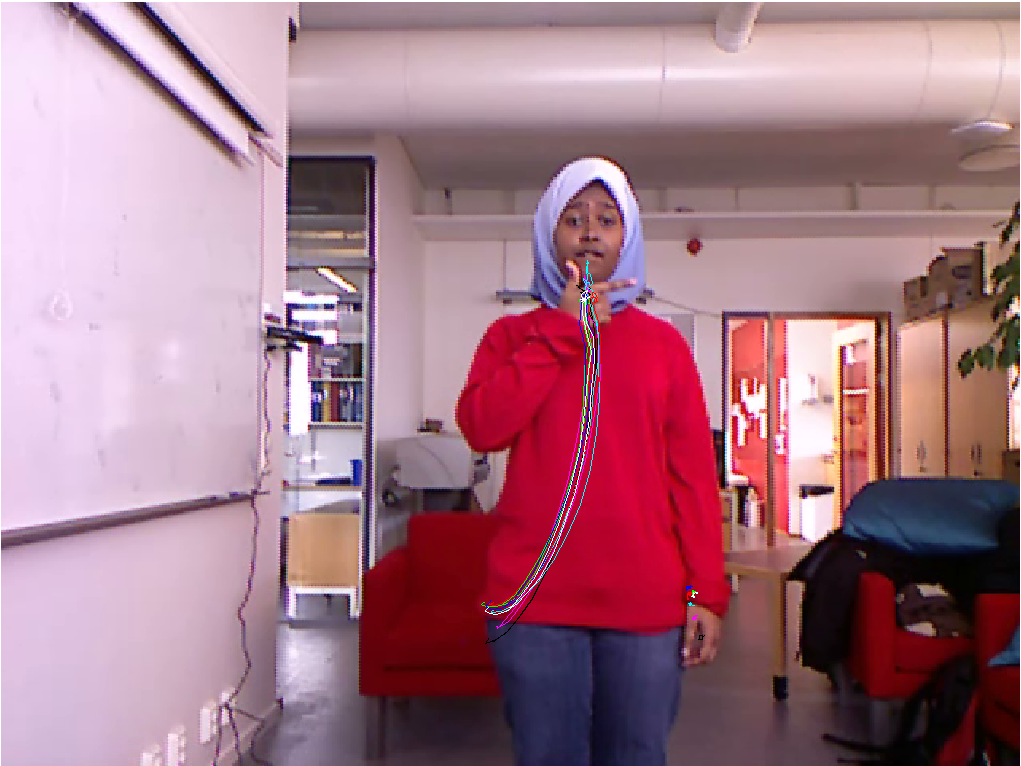}}}
  \caption{Automatic sign language recognition (SLR) from video is a challenging task, due to both high style variation between signers, and very subtle differences between different signs.  (a) A sign with high intra-class variation. This is an example of a sign which some signers perform with one hand, and others with two. Moreover, there are style differences between the two signers. (The signer to the left is left handed, something that can be addressed easily by mirroring the video.) (b-c) This is an example of two signs which are difficult to distinguish using hand shape and pose only.}
  \label{fig:si1}
\end{figure}

A challenge for SLR systems is also the fact that the same sign can appear differently depending on who is performing it. A signer independent SLR method needs to handle these differences. Fig.~\ref{fig:si1}(a) shows two performances of the sign 'Kan jag f\aa' ('Can I have'). Some signers perform this sign with two hands while others do it with one hand. This can be handled by creating multiple classes for each sign. In addition to that, different signers have different signing styles (how they move their hand and how they make the characteristic hand shapes). Thus, certain aspects of the signing reflect the class of sign being signed, while other aspects reflect the individual style of the signer. An ideal signer independent classifier would ignore the individual style aspects and focus only on class-relevant aspects. 

At the same time, two different signs can display very similar visual features, as exemplified in Fig.~\ref{fig:si1}(b-c). Thus, it is important to use as descriptive features as possible. This is a trade-off -- low dimensional and signer independent features are robust to intra-class style differences, while high dimensional and rich features are able to pick up on subtle inter-class dependencies.

This paper gives four contributions. Firstly, we provide a {\em method} -- to our knowledge the first one in the literature -- {\em for automatic recognition of} a challenging set of {\em Swedish Sign Language (SSL)} words. Secondly, we introduce a {\em robust hand segmentation method which employs RGB-D (Kinect) video}. Third, and foremost, we use Fisher Linear Discriminant Analysis (LDA) \cite{duda00} to {\em find the most discriminative directions in the feature space across multiple signers}. Finally, we perform sign language recognition from 3D hand trajectories; we {\em show that the introduction of depth into the hand motion representation increases the recognition rate}.

\subsection{Application: Supportive Signing}

The deaf population is not the only group that use signing to communicate. There is a large group of people who use verbal communication but rely on signing as a complement. Children born with hearing impairment or some form of communication disability such as developmental disorder, language disorder, cerebral palsy or autism, frequently have the need for this type of augmented and reinforced communication. These communication forms are known as TSS (``Signs as Support") in Sweden. They function by "borrowing" individual signs from a signed language (e.g., TSS borrows from SSL). The borrowed signs support and enforce the verbal communication. As such, these communication support schemes do away with the grammatical constructs in sign language and keep only parts of the vocabulary. 
 
One important difference between SSL and TSS is that the latter is poorly formalized and described, and the extent and manner in which it is taught differ widely between different parts of the country. While many deaf children have sign language as their first language and are able to pick it up in a natural way from the environment, children that need signs for other reasons do not have the same rights and opportunities to be introduced to signs and signing. The Swedish TIVOLI project aims at creating a learning environment where children can pick up signs in a game-like setting. An on-screen avatar presents the signs and gives the child certain tasks to accomplish, and in doing so the child gets to practice the signs. The system is thus required to interpret the signs produced by the child and distinguish them from other signs, and indicate whether or not it is the right one and if it was properly carried out. This is a very challenging task due to the large variability that is expected. The sign recognition module should be able to cope with difference in environment, lighting, subject clothing, and subject size.  Due to the nature of supportive signing, the system only has to consider the base forms of isolated signs.

The method presented here will serve as the key recognition component of the system. Section \ref{sec:handseg} presents the hand segmentation and feature extraction from RGB-D video, Section \ref{sec:LDA} explains how signer independent features are learned and Section \ref{sec:classification} outlines the classification of signs represented by the extracted features. Experiments in Section \ref{sec:experiments} show the method to distinguish signs from a challenging 94 word vocabulary with a precision of 86\% on average, 94\% for the most skilled signer, when the method was trained and tested on a single signer, and 30\% on average, 46\% for one signer, when trained on multiple signers and tested on a new signer. Moreover, the introduction of depth into the hand motion representation increases the recognition rate with 10\% in the signer dependent case and 25\% in the signer independent case.

\section{RELATED WORK}
\label{sec:related}

We here only focus on non-intrusive video based automatic sign language recognition for dynamic signs; more comprehensive reviews can be found in \cite{s1} and \cite{s5}. Methods using intrusive data gloves were common at the start of sign language recognition research but in the last decade, vision based methods have become more common and they have started tackling difficult problems like, large vocabularies \cite{41} and sign language recognition in uncontrolled environment \cite{5,33}. It is common for vision based methods to restrict the background (uniformly colored or static), require the signer to wear full sleeved clothing and even in some cases to wear colored gloves. These restrictions make the task of hand segmentation and tracking significantly easier but at a cost of limiting the system's usability. 

Vision based methods rely on multiple image cues to detect and segment hands, these cues include color, motion \cite{1} (frame differencing), edges, background subtraction and region context. Both statistical and adaptive skin color models \cite{2} are common. The adaptive color models can adapt to take different environmental condition (e.g. illumination, signer) in account by changing their model parameters usually using first few or few recent frames in the video sequence.

In the domain of sign language, Kalman filters \cite{1,2} are the most common method used for tracking hands, used in this paper. \cite{5} used a multiple hypothesis approach and chose the most likely hand trajectory at the end of a sign. Other hand tracking approaches include dynamic programming \cite{28}.

Non-manual features, such as head pose and motion \cite{10,11}, facial expression \cite{12}, gaze \cite{52}, and lip shape \cite{11}, convey very useful information in sign language. In the last decade, facial expressions have received the most attention among the non-manual features.

The majority of systems only support signer dependent operation, i.e., every user is required to train the system before being able to use it.  Signer independence is usually implemented by some normalization of features, e.g., with respect to the body proportions of the signer \cite{5}, or by parametrizing the model by signer identity \cite{13}, or training the model with multiple signers \cite{5}, an approach that works well if the features are robust \cite{33}. The down-side of robust, crude features are that the precision is lower, which makes the classification task harder.

Most early sign language recognition systems used a form of template matching or neural networks for recognition \cite{14}. However in the last decade or two, Hidden Markov Models (HMM) \cite{hmm89} have become the most common classification method for SLR. Some of the other common methods include Conditional Random Fields (CRF) \cite{4}, Dynamic Time Warping (DTW) \cite{31} and nearest neighbor.

\section{HAND SEGMENTATION AND FEATURE EXTRACTION}
\label{sec:handseg}

\begin{figure*}[t]
\centerline{%
\subfigure[Hands close to body]{\includegraphics[width=0.32\textwidth]{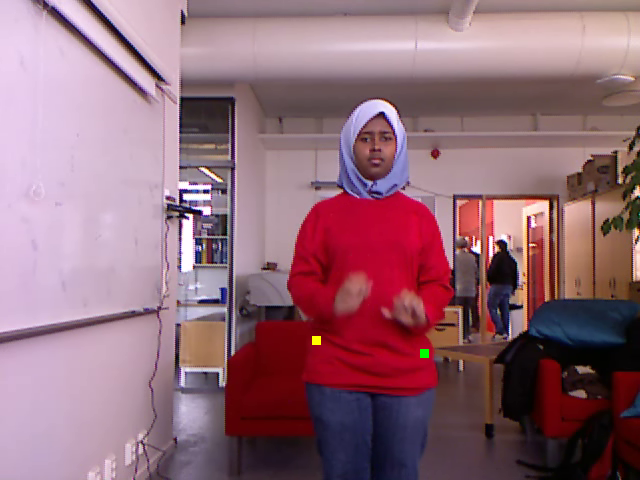}} ~
\subfigure[Hands moving fast]{\includegraphics[width=0.32\textwidth]{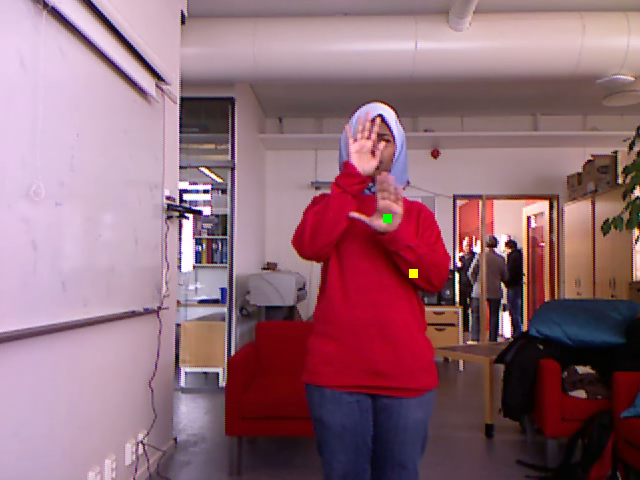}} ~
\subfigure[Hands occluding shoulders]{\includegraphics[width=0.32\textwidth]{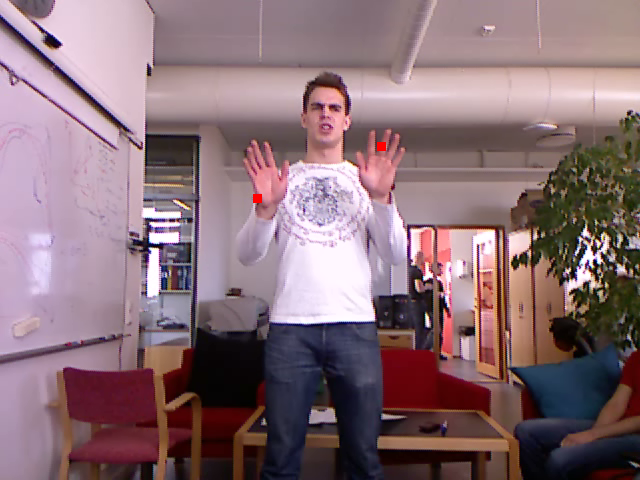}}}
\caption{Examples of error modes in the Kinect tracker. Right (\textcolor{yellow}{$\bullet$}) and left (\textcolor{green}{$\bullet$}) hand, and shoulder (\textcolor{red}{$\bullet$}) pose estimates are marked.}
\label{fig:kinectError}
\end{figure*}

The input to the method is RGB-D (Kinect) video of the signer, along with a 3D estimation of the signer's skeleton from the Microsoft Kinect tracker \cite{Shotton2011}. 

Designed for gaming, the Kinect tracker provides the position of hands but this position is not accurate enough to restrict the search space for hands, as can be seen in Fig.~\ref{fig:kinectError}(a-b). Therefore, a separate hand tracking method (described below) is used to capture hand pose, while the head and torso pose are captured with the built-in tracker.

\subsection{Skin Detection}
The signer is required to wear full sleeved, non-skin colored clothing to simplify the hand segmentation. The search space for each hand is limited to a small rectangular region in each frame, centered in the Kalman hand pose estimate. Initially the signer is segmented from the background using the Kinect skeleton information. 

Human skin color has a restricted range of hue and saturation but there is significant variation in luminance. We thus represent color in terms of a normalized RGB colorspace (($r = \frac{R}{(R+G+B)}$ and $g = \frac{G}{(R+G+B)}$)). Two histogram models (one for skin color and another for non-skin color) are trained using the Compaq dataset \cite{Compaq} These general models are applied to the first frame of each sign in order to get an initial segmentation mask. Skin and non-skin training data, specific to the current signer, are extracted from this mask. These data are used to initialize adaptive, signer specific histogram models, which are in turn updated at each time step $t$ as a linear combination of the models from $t-1$ and $t$. 

The output in each time step $t$ is a binary skin mask $S_t$. 




Another cue to improve the skin detection performance is to use pixel change value. Since among the signer pixels only the hand and arm move frequently, rest of the body remain relatively static. This information can be used to narrow down the pixels which can potentially belong to the hands in any frame. Any pixel with significant change in its value has a higher chance of belonging to the hands. A motion change measure $P^{M}_t$ is created by taking the image difference of current and previous grey-scale image. 

$P^{M}_t$ is thresholded to a binary motion mask $M_{t}$.


The resulting segmentation mask is defined as $S_{t}\,\mathrm{AND}\,M_{t}$. Then morphological operations (erosion followed by dilation) are applied to remove spurious regions. The resulting potential hand segments are ranked based on their depth (distance from camera), size, and distance from the predicted hand position. The segments with the highest score are assigned to the hands.

\subsection{Hand Tracker}

Two Kalman filters are used for each hand; one to keep track of the position of hands ($x$ and $y$), velocity ($\dot{x}$ and $\dot{y}$), acceleration ($\ddot{x}$ and $\ddot{y}$), the other keeps track of the width and height of the bounding box ($w$ and $h$) around each hand, and their rate of change ($\dot{w}$ and $\dot{h}$). Some extra padding is added to all four sides of both bounding boxes to accommodate errors in position estimate (when hands change their movement direction or shape abruptly). When hands overlap, this event is detected, and they are treated as one hand. 

\subsection{Occlusion Handling}
When the predicted bounding box for both hands overlap and there is only one large skin object in the hand search space, hand over hand occlusion's start is marked. During most of the hand over hand occlusion, both hands are either touching each other or are very close to each other. This makes the task of recovering hand shape accurately even with depth information very difficult. Since hand shape usually remains almost same over few consecutive frames in sign language, the hand shape from last frame before occlusion started is used to locate the position of each hand in the joint blob using 
template matching. 
During the occlusion, only the position of hands is updated, hand shape features are retained from the last frame before occlusion started.

\begin{figure*}[t]
\centering
\includegraphics[width=\textwidth]{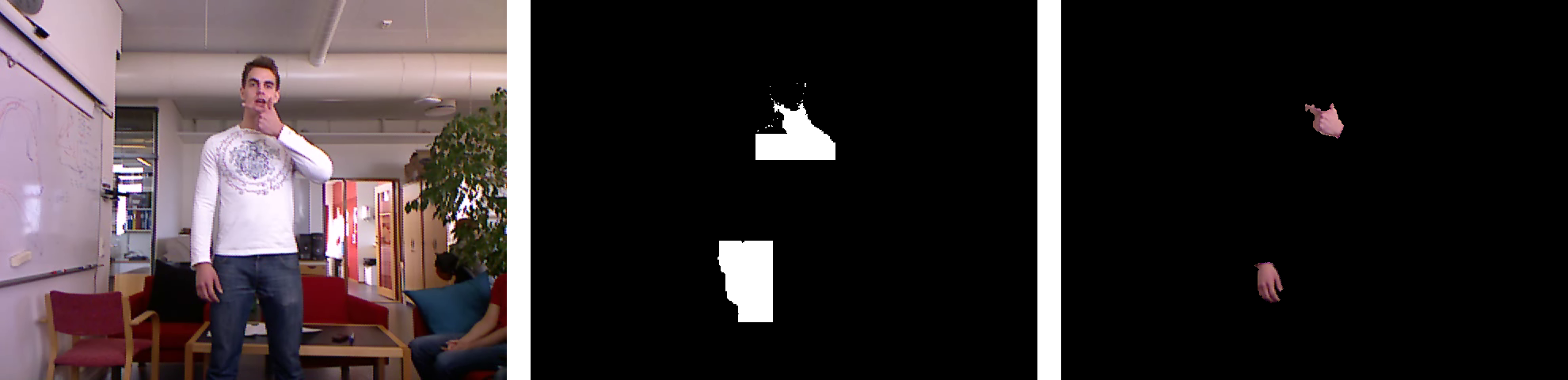}\\[2mm]
\includegraphics[width=\textwidth]{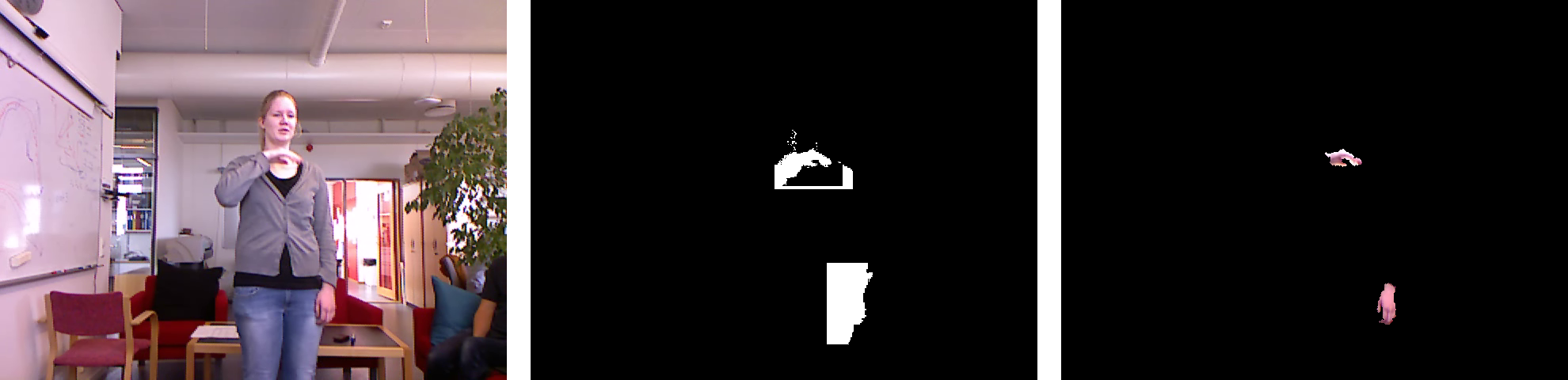}
\caption{Two examples of hand segmentation. (left) Frame. (middle) Hands after initial segmentation and face removal. (right) Segmented hands. {\em For more examples, see the video in the Supplementary Material.}}
\label{fig:seg}
\end{figure*}

Another common occlusion in sign language is hand over face occlusion. This occlusion is detected when one of the hand bounding box overlaps the face bounding box. This project solves this type of occlusion using the depth information. A depth model (depth of each pixel) in the face bounding box is created at the start of each sign. This model is updated at each frame. When this occlusion starts, current depth at each pixels is subtracted from depth of each pixel in face model to find out the foreground (hand) pixels. These foreground pixels become the part of the search space for hands. While the occlusion lasts, depth of all pixels excluding those that belong to the hand are updated normally. Fig.~\ref{fig:seg} shows two frames with hand over face occlusion along with search space for hands after background and face removal and the final segmented hands.

\subsection{Features}
The following features are extracted from each frame:
\begin{itemize}
\item Image position ($x$ and $y$) and velocity ($\dot{x}$ and $\dot{y}$) of each hand centroid relative to neck,
\item Depth position ($z$) and velocity ($\dot{z}$) of each hand centroid relative to torso,
\item Area ($a$), the number of pixels in the segmented hand which is robust to segmentation errors, and perimeter ($p$), the number of pixels around the border of the segment which is sensitive to segmentation errors, 
\item Solidity ($s$), the proportion of the pixels in the convex hull that are also in the hand segment,
\item Major ($M$) and minor ($m$) axis length of the ellipse that has the same normalized second central moments as the hand blob, measuring the size of this ellipse,
\item Eccentricity ($c = \sqrt{(1 - \frac{m}{M})^2}$) of the ellipse, measuring how much the hand blob deviates from circular,
\item Angle ($\cos(\beta)$) between the major ellipse axis and the horizontal image axis,
\item Hu invariant moments \cite{hu} ($\mathbf{HU}$) which are invariant to image scale, translation and rotation.
\item Shape Context \cite{SC} ($\mathbf{SC}$). 40 points on the boundary of each hand are chosen, distance and direction to other points is calculated in log-polar space. 5 bins are used for distance (normalized using median distance) and 9 bins for orientation.
\item HOG \cite{hog} ($\mathbf{HOG}$). Each hand is divided into 2$\times$2 cells and orientation is binned into 9 bin histograms.

\end{itemize}

Position features are calculated relative to the neck joint of the signer, neck is chosen because it is one of the most stable and accurately tracked joints in the upper torso. Position, perimeter, velocity, and major and minor axes are then normalized using the distance between the signer's shoulders. Distance between shoulders is calculated using the first few frames of each sign, when hands are not likely to be occluding shoulders, something that makes the Kinect tracker compute erroneous shoulder positions, see Fig.~\ref{fig:kinectError}(c). Area is normalized using the square of the distance between the signer's shoulders. Features for idle hand (the hand which does not move during the whole sign) are set to 0 in a pre-processing step.

We studied the following feature sets (Section \ref{sec:experiments}):
\begin{itemize}
	\item $\mathbf{posXYZ} = (x, y, z)$
	\item $\mathbf{posXY} = (x, y)$ from Kalman Tracker
	\item $\mathbf{posKinect} = (x, y)$ from Microsoft Kinect Tracker
	\item $\mathbf{velocityXYZ} = (\dot{x}, \dot{y}, \dot{z})$
	\item $\mathbf{pos} = (x, y, z, \dot{x}, \dot{y}, \dot{z})$	
	\item $\mathbf{S} = (a, p, s, c, M, m, \cos(\beta))$	
\end{itemize}

\section{LEARNING SIGNER INDEPENDENT FEATURES}
\label{sec:LDA}

\begin{figure}[t]
\centerline{%
\subfigure['Smaka'  ('Taste')]{\includegraphics[width=0.24\textwidth]{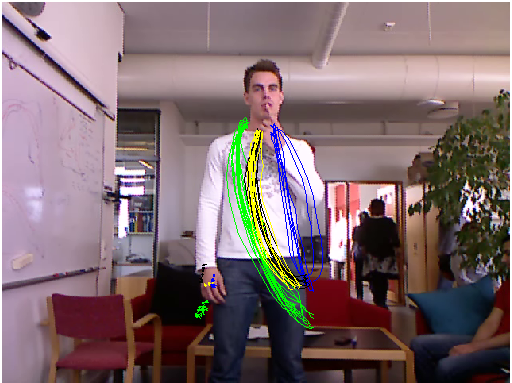}} 
\subfigure['5']{\includegraphics[width=0.24\textwidth]{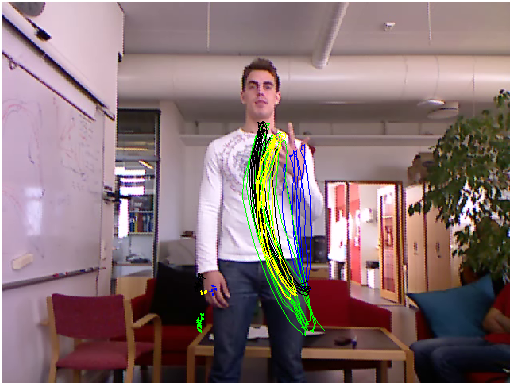}}}
\caption{Trajectories for all 4 signers for signs 'Smaka'  ('Taste') and '5' : Signer A ($\bullet$), Signer B (\textcolor{blue}{$\bullet$}), Signer C (\textcolor{yellow}{$\bullet$}), Signer D (\textcolor{green}{$\bullet$})}
\label{fig:overlappedTraj}
\end{figure}

There is considerable variation between the signing style of different signers as illustrated in Fig.~\ref{fig:overlappedTraj}. This causes some features (e.g., the horizontal position as in Fig.~\ref{fig:overlappedTraj}) to vary significantly across signers. HMMs learn the variance of all features for each specific sign, independently of other signs. In the case of Fig.~\ref{fig:overlappedTraj}, the variance over horizontal position for the signs 'Smaka'  ('Taste') and '5'  will be large in the HMM states -- the horizontal position will then not be taken into regard very much in evaluating a new sign using this model.
However, imagine that there is another sign which only deviates from 'Smaka' in terms of horizontal position. Then, the 'Smaka' classifier will give an instance of that sign a high probability of being 'Smaka'.  

In contrast, the influence of different features should be decreased or increased based on how discriminative the feature is in separating that sign from other signs. Using Fisher Linear Discriminant Analysis (LDA) \cite{duda00} -- more precisely, its multi-class equivalent -- we transform our features to a new feature space, in which a few dimensions contain most of the discriminatory information. This results in selection of features which are more discriminating across multiple signers and reduces the impact of individual signing style of signers.

Initially feature vectors have to be aligned because the actual signs in different samples $s$ have different start and stop frames. Since we are looking at the difference between different samples at frame level, it is required that all frames within all samples are aligned. We use Dynamic Time Warping (DTW) to achieve this, alignment is done using only position features $\mathbf{posXY}$. 

Once all the signs are aligned, they are re-sampled so that they have equal length. The feature vectors from first and last third of the frames are discarded because the motion and hand shape at the start and end of the signs is similar and the middle portion $X$ is most likely to contain the information particular to each sign. At each frame $t$, the means $\mu^{c}_{t}$ of each sign $c$ and the total mean $\mu_{t}$ are calculated as 
\begin{equation}
\label{eq:meanc}
	\mu^{c}_{t} = \frac{1}{N^c}\sum_{n=1}^{N^{c}} X^{c}_{t,n}
\end{equation}
\begin{equation}
\label{eq:mean}
	\mu_{t} = \sum_{c=1}^{C} \frac{N^{c}}{\sum_{c=1}^{C} N^{c} } \mu^{c}_{t}
\end{equation}
where $N^{c}$ is the number of samples of class $c$, and $X^{c}_{t,n}$ is frame $t$ of sample number $n$ from sign class $c$. 

These mean values from all frames are combined to get the total between-sign scatter matrix $S^B$ and within-sign scatter matrix $S^W$ as 
\begin{equation}
\label{eq:sb}
	S^B = \sum_{t=1}^{T} \sum_{c=1}^{C} \left(\mu^{c}_{t} - \mu_{t}  \right)  \left(\mu^{c}_{t} - \mu_{t}  \right)^{T}
\end{equation}
\begin{equation}
\label{eq:sw}
	S^W = \sum_{t=1}^{T} \sum_{c=1}^{C} \left( \frac{N^{c}}{\sum_{c=1}^{C} N^{c} } \right) \sum_{n=1}^{N^{c}} \left( X^{c}_{t,n} - \mu^{c}_{t}  \right)  \left( X^{c}_{t,n} - \mu^{c}_{t}  \right)^{T}
\end{equation}
 
These scatter matrices are used to find the eigenbasis $W$ that maximizes the difference between different signs while minimizing the difference between different samples of same sign. This transformation can be found by solving the characteristic polynomial
\begin{equation}
\label{eq:lda}
	| S^B - \lambda_i S^W | = 0
\end{equation}
to get the eigenvalues $\lambda_i$, and then solving for each eigenvector $W_i$ the equation
\begin{equation}
\label{eq:lda}
	(S^B - \lambda_i S^W ) W_i = 0
\end{equation}

The $M$ eigenvectors $W_i$ with the highest eigenvalues $\lambda_i$ are selected, and a signer-independent representation of the features is obtained by projecting them onto this new $M$-dimensional feature space.

\section{Sign Classification}
\label{sec:classification}

HMM \cite{hmm89}, which are very common in sign language recognition, were used to represent each sign as a sequence of states in the feature space. 
In our implementation we used a chain of $N=7$ active (emitting) states and two (first and last state) non-emitting states, with no skip states. The number of states were determined empirically. The HMM models were trained using the Georgia Tech Gesture Toolkit \cite{gt2k}.


\section{EXPERIMENTS}
\label{sec:experiments}

\subsection{Data collection.}
We have collected our own data since our method requires RGB-D video of Swedish Sign Language, which is not available in any of the existing corpora. The vocabulary for this project consists of 94 words, divided into 4 groups ranging in size from 26 to 20. Three of the groups correspond to different games in the tutor system described in the Introduction: {\em Horror House}, {\em Fun House}, and {\em Ice-Cream Stand}, while the fourth, {\em Numbers}, consists of the numbers 1 to 20. The group {\em Numbers} consist of many signs which are static and differ from each other only in the number of opened hand fingers. 

Some signs are extremely similar in terms of hand motion. An example is shown in Figure \ref{fig:si1}(b-c):  The only difference between the two signs is the facial expression, which indicates that the tutoring application would benefit from the inclusion of non-manual features in the method (see the Conclusions).

Data (two videos and one text file for each signer) were recorded using Microsoft Kinect and Kinect for Windows SDK. Both depth and color videos had a resolution of $640\times480$ and a frame rate of $30$ Hz. The text file contains the location of all joints in the upper body of signer tracked by Kinect. The dataset consists of a total of 23 samples of each sign performed by 4 signers. Three signers (A, C and D) were female and right-handed, while the fourth signer (B) was left-handed and male. For the left handed signer, images and joint positions were flipped horizontally. The number of samples (where each sample contains one instance of each of the 94 signs) recorded were 7 for signer A, 4 for signer B, 5 for signer C and 7 for signer D. 


\begin{table*}[t]
\caption{Comparison of different shape features}
\label{tab:exShape}
\centerline{\begin{tabular}{| c | c | c | c | c | c || c | c | c | c | c |}
  \hline
    & \multicolumn{5}{| c ||}{Signer Dependent} & \multicolumn{5}{| c |}{Signer Independent} \\ \hline
	& mean & A 	& B 	& C 	& D & mean & A 	& B 	& C 	& D \\ \hline
  \small  $\mathbf{S}$ 			& 77.65\%  & 84.95\% & 68.88\% & 78.51\% & 78.27\% & 14.14\%  & 14.13\% & 14.63\% & 17.02\% & 10.79\% \\  \hline
  \small  $\mathbf{HOG}$ 		& 83.36\%  & 88.75\% & 72.34\% & 85.11\% & 87.23\% & 17.32\%  & 21.73\% & 9.31\% & 22.13\% & 16.11\% \\   \hline
  \small  $\mathbf{SC}$			& 57.02\%  & 67.48\% & 45.48\% & 61.49\% & 53.65\% & 15.15\%  & 19.15\% & 13.56\% & 15.74\% & 12.16\% \\   \hline
  \small  $\mathbf{HU}$ 			& 26.95\%  & 25.84\% & 21.28\% & 34.26\% & 26.44\% & 5.70\%  & 7.60\% & 6.12\% & 4.68\% & 4.41\% \\  \hline
  \end{tabular}}
\end{table*}

All experiments are done with leave one out cross-validation; in the signer dependent case, the classifier is trained on all samples but one for this signer, and tested on the last sample, and in the signer independent case, the classifier is trained on three of the signers and tested on the fourth one.

Setting idle hand features to zero results in significant improvement in the recognition rate when using only shape, or shape and position features and some minor improvement when using position features. This is because it removes the impact of these features which are not needed for recognition but can still significantly lower the probability of a HMM model generating the observed sequence during testing. Min-Max normalization was attempted but it resulted in slightly lower recognition rate compared to not-normalized data and as features are in the same value range, no normalization was used in the experiments.

\subsection{Feature Selection}
\label{sec:exFeature}

\subsubsection{Shape Features}
\label{sec:exShape}

Table \ref{tab:exShape} presents the recognition rate for various features used to represent hand shape. The performance of $\mathbf{HOG}$ features is better than all other shape features in both signer dependent (83.36\%) and signer independent (17.32\%) tests. One reason is that the HOG descriptor takes the entire hand pattern into account, while the other features are extracted from the silhouette only. Another reason might be that HOGs are not strongly impacted by minor segmentation errors. 

Some features in $\mathbf{S}$ are also quite robust to segmentation errors and slight changes in hand shape, which is why it performs better than $\mathbf{HU}$ which is sensitive to these sources of noise.  The reason for the weak performance of $\mathbf{SC}$ is probably low image resolution; the hand size in many frames is very small, which means that a small segmentation error can result in a significantly different shape context. 

The performance of  $\mathbf{S}$ and $\mathbf{HOG}$ improves when used in combination with positional features. This is discussed in Section \ref{sec:signerdep}.

\subsubsection{Position Features}
\begin{table}[b]
\caption{Comparison of position features}
\label{tab:exPos}
\centerline{\begin{tabular}{| c | c | c | c | c | c |}
  \hline
   & mean & A & B & C & D \\
  \hline
  \scriptsize  $\mathbf{pos}$ (SD) & 71.16\%  &  75.08\% & 70.21\% &  67.45\% & 71.88\% \\   \hline  
  \scriptsize  $\mathbf{posXYZ}$ (SD) & 67.86\% & 73.10\%  & 65.69\% & 65.32\% & 67.33\% \\  \hline
  \scriptsize  $\mathbf{posXY}$ (SD) & 61.83\% & 69.00\%  & 59.31\% & 58.09\% & 60.94\% \\   \hline
  \scriptsize  $\mathbf{posKinect}$ (SD) & 52.84\% & 52.13\%  & 50.00\% & - & 56.38\% \\   \hline
  \scriptsize  $\mathbf{pos}$ (SI) & 17.47\% & 22.80\% &  5.05\% &  32.77\% & 9.27\% \\   \hline  
  \scriptsize  $\mathbf{posXYZ}$ (SI) & 13.83\% & 18.84\%  & 5.05\% & 28.09\% & 3.34\% \\  \hline
  \scriptsize  $\mathbf{posXY}$ (SI) & 13.79\% & 14.74\%  &  9.31\% & 27.02\% &  4.10\% \\   \hline  
\end{tabular}}
\end{table}

In this experiment, the aim is to evaluate the amount of information contained in the different pose and velocity features. idle hand features were not set to zero. 

Since there is no ground truth of hand positions available, our Kalman tracker was compared to the Kinect tracker by training classifiers with both these sets of features.  As can be seen in Table \ref{tab:exPos}, the position estimate from our Kalman tracker gives 17\% higher recognition accuracy on average, than the same position estimates from the built-in Kinect tracker. 

Introducing depth $z$ resulted in an improvement of 10\% compared to when only $x$ and $y$ were used in the case of signer dependent recognition and almost same performance in the case of signer independent recognition. When hand shape features are used in addition to position features, the improvement is small; however, this indicates that depth adds robustness to the classifier when color based segmentation fails or can not be attempted (half sleeved clothing, etc).

The signer independent recognition accuracy for signer B and D is significantly lower than signer A and C for all feature combinations. The reason for this is that the trajectories of these two signers deviates in a systematic manner from other signers, as can be seen in Fig.~\ref{fig:overlappedTraj}. These signers also have higher variance in their trajectories across different samples of same sign. Both these signers were learners of sign language, which can explain the high variability.

\subsection{Signer Dependent Recognition}
\label{sec:signerdep}
\begin{table}[b]
\caption{Accuracy of signer dependent recognition}
\label{tab:exSD}
\centerline{\begin{tabular}{| c | c | c | c | c | c |}
  \hline
    & mean & A & B & C & D \\
  \hline
  \scriptsize $(\mathbf{pos}, \mathbf{S})$  & 85.06\% & 91.19\%  & 77.39\% & 83.19\% & 88.45\% \\ \hline
  \scriptsize $(\mathbf{pos}, \mathbf{HOG})$  & 85.45\% & 91.95\%  & 74.73\% & 87.45\% & 87.69\% \\ \hline
  \scriptsize $(\mathbf{pos}, \mathbf{S}, \mathbf{HOG})$  & 87.05\% & 93.62\%  & 77.13\% & 87.02\% & 90.43\% \\
   \hline
\end{tabular}}
\end{table}

In this experiment, separate HMM models for signers A, B, C, and D were trained and tested using the $\mathbf{HOG}$,  $\mathbf{pos}$, and  $\mathbf{S}$ feature sets, judged in the previous experiments to perform the best. For each signer, the training was performed on all samples but one, and testing was carried out with the left out sample. All combinations of training and testing samples for each signer were evaluated, and the reported results for each signer A, B, C, and D are the mean of these recognition rates. 

Table \ref{tab:exSD} lists the recognition rate for all four signers. First of all, it is evident that the feature sets $\mathbf{HOG}$ and  $\mathbf{S}$ carry complementary information; the recognition rate improves consistently with 2-3\% when they are combined. This makes sense since $\mathbf{S}$ contains information about the area and perimeter of the hand, while $\mathbf{HOG}$ contains complementary information about the shape of the hand, irrespective of absolute scaling.

When studying the recognition using $(\mathbf{pos}, \mathbf{S}, \mathbf{HOG})$, signer A had the highest recognition rate of 93.62\% while signer B had the lowest recognition rate of 77.13\%. 

The reasons for the good recognition of signer A are most probably that she is the most skilled signer of the four, but also that her clothing color is significantly different from her skin color, which gives a very accurate skin segmentation. Fig. \ref{fig:exSDConfusionMatrix}(a) shows the confusion matrix for signer A. The recognizer confuses the last 20 signs to a higher degree than the first 74. These signs are from the group 'Numbers', which contains signs with very similar hand shapes. Some of the other signs with low recognition rate have another sign with very similar hand shape and trajectory in the vocabulary, e.g. Fig.~\ref{fig:si1}(b-c), which only differ from each other in their facial expression. 

One of the reasons why signer B has lower recognition rate than other signers is that there were only 3 training samples for this signer.

\subsection{Signer Independent Recognition after LDA Feature Transformation}

\begin{table}[b!]
\caption{Accuracy of signer independent recognition after LDA feature transformation (number in () lists \# dimensions used)}
\label{tab:exLDA}
\centerline{\begin{tabular}{| c | c | c | c | c |}
  \hline
     & \multicolumn{4}{ c |}{$(\mathbf{pos}, \mathbf{S})$}\\ \hline  
			& no LDA(26) & LDA(15) & LDA(20) & LDA(25) \\ \hline
	mean	&	29.04\% & 34.30\% &35.11\% &34.94\%  \\ \hline
	A		&	32.37\% & 41.64\% &42.86\% &43.92\%  \\ \hline
	B		&	16.76\% & 16.76\% &18.09\% &16.22\%  \\ \hline
	C		&	42.55\% & 46.60\% &45.74\% &45.74\%  \\ \hline
	D		&	24.47\% & 32.22\% &33.74\% &33.89\%  \\ \hline\hline
     & \multicolumn{4}{ c |}{$(\mathbf{pos}, \mathbf{HOG})$}\\ \hline  
			& no LDA(84)  & LDA(20) & LDA(25) & LDA(30)  \\ \hline
	mean	 &	23.00\% &27.41\% &	26.66\% & 27.28\%  \\ \hline
	A		&	27.05\% &37.84\% &	33.43\% & 34.50\%  \\ \hline
	B		&	13.30\% & 3.72\% &	4.79\% & 5.85\%  \\ \hline
	C		&	29.15\% &37.66\% &	35.74\% & 34.89\%  \\ \hline
	D		&	22.49\% &30.40\% &	32.67\% & 33.89\%  \\ \hline\hline
     & \multicolumn{4}{ c |}{$(\mathbf{pos}, \mathbf{S}, \mathbf{HOG})$}\\ \hline  
			& no LDA(98) & LDA(20) & LDA(25) & LDA(30)  \\ \hline
	mean	&	27.12\% &	31.40\% &	31.84\% & 30.47\% \\ \hline
	A		&	29.94\% &	36.78\% &	39.97\% & 36.32\% \\ \hline
	B		&	17.82\% &	4.52\% &	4.52\% & 6.12\% \\ \hline
	C		&	34.89\% &	45.11\% &	44.89\% & 41.91\% \\ \hline
	D		&	25.84\% &	39.21\% &	37.99\% & 37.54\% \\ \hline
\end{tabular}}
\end{table}

\begin{figure*}[t]
\centering
\subfigure[Signer dependent (93.62\%)]{\includegraphics[width=0.32\textwidth]{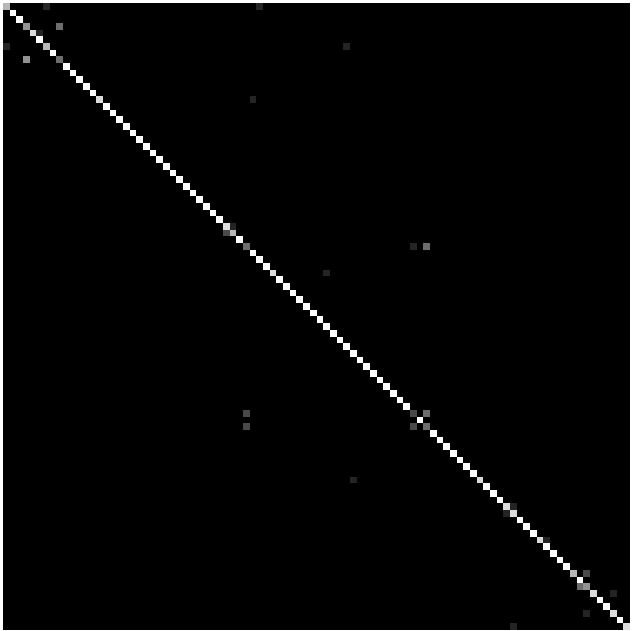}} ~
\subfigure[Signer independent, no LDA (36.78\%)]{\includegraphics[width=0.32\textwidth]{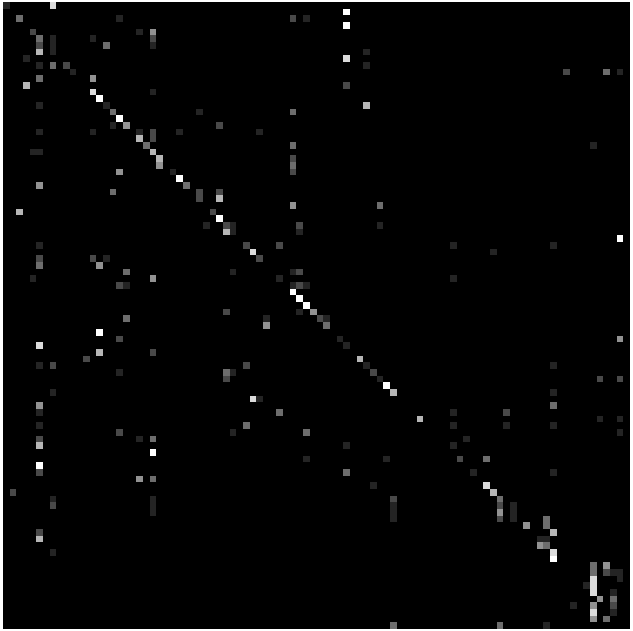}} ~
\subfigure[Signer independent (39.97\%), LDA(25)]{\includegraphics[width=0.32\textwidth]{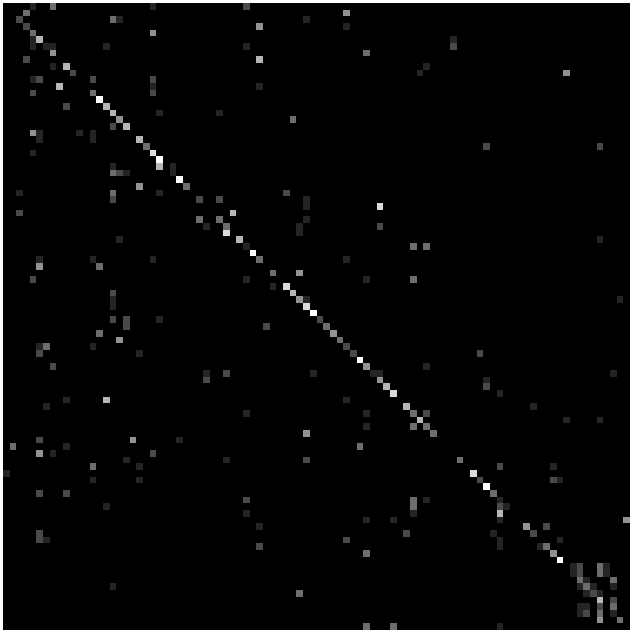}}
\caption{Confusion matrices for signer A, using features $(\mathbf{pos}, \mathbf{S}, \mathbf{HOG})$.}
\label{fig:exSDConfusionMatrix}
\end{figure*}

In this experiment, 3 signers were used to learn the LDA feature transformation $W$, described in Section \ref{sec:LDA}. The learned transformation was applied to features from these signers before they were used for training HMM models. When testing, features from the left out signer were transformed using $W$, and classified using the trained HMM. This was repeated four times, leaving signer A, B, C, and D out for testing in turn. The results are listed in Table \ref{tab:exLDA}, and compared with a baseline of performing the training without prior pre-processing using LDA. 

The mean improvement of LDA is 10-15\%. For signer A, C and D, the increase in performance was even higher, 8-37\%. The reason for this is probably that the signing styles of these signers deviated from each other in predictable ways -- they have all developed an individual, consistent style, just as one develops an individual style for hand writing. However, LDA did not improve the recognition of signer B. The reason is, in analogy to above, that signer B has not yet developed a consistent signing style since he is a learner; his signs vary in a stochastic manner, and the patterns of style variation learned from the other signers do not apply on his signing. 

Looking at the confusion matrices in Fig.~\ref{fig:exSDConfusionMatrix}(b) and (c), it is evident that the recognition of some signs has greatly improved by the LDA preprocessing -- the signs that show a great style variability between signers -- while others are unaffected -- the signs that are performed in the same manner independent of signer, or with more random variations.

\section{CONCLUSIONS}
\label{sec:conclusions}

We present a method for recognizing Swedish Sign Language (SSL) from video. The method will be used in a computer game intended for sign language training for children with communicative disabilities. The signer is captured with an RGB-D (Kinect) system, which gives the possibility to recognize signs in terms of 3D hand motion. The primary contribution of this paper is a method to learn a projection of the hand features using Linear Discriminant Analysis (LDA), which makes the recognition robust to variation between different signers. Additional contributions of the paper are II) the SSL recognition method, III) a robust, hand segmentation method based on color, depth and motion, and IV) the inclusion of hand depth among the classification features.

Since there is no commonly used dataset for sign language recognition, it is very difficult to compare different methods quantitatively. Differences in vocabulary across papers is also very varied, which further complicates the comparison. \cite{5} achieved up to 99\% recognition accuracy for signer dependent and 44\% in signer independent experiments in controlled environments. Our system has a recognition accuracy of 87\% on average in signer dependent experiments and 35\% on average in signer independent experiments in a natural environment with a very demanding set of signs.

The present method can primarily be extended in three directions. Firstly, as illustrated in Fig.~\ref{fig:si1}(b-c), certain signs can not be distinguished by manual features alone. We will investigate the inclusion of non-manual features in the recognition. 

Secondly, LDA is a linear method for finding the most discriminative directions in feature space. We will investigate non-linear alternatives, such as Gaussian Process Latent Variable Models \cite{lawrence04}. This can be expected to improve performance, as the Gaussian assumption on data distribution posed by LDA is probably a simplification of the real data distribution.

Moreover, system will be tested using the data from children with communication disabilities.

\bibliographystyle{IEEEtran}
\bibliography{IEEEabrv,FG2013}
\end{document}